\documentclass[twocolumn]{article}
\usepackage{spconf,amsmath,epsfig}
\usepackage{multirow}
\usepackage{multicol}
\usepackage{float}
\usepackage{colortbl}
\usepackage{booktabs}
\usepackage{amssymb}
\usepackage{bm}
\usepackage{afterpage}
\usepackage{caption}
\usepackage{graphicx}
\usepackage{bm}
\usepackage[pagebackref,breaklinks,colorlinks]{hyperref}

\let\OLDthebibliography\thebibliography
\renewcommand\thebibliography[1]{
  \OLDthebibliography{#1}
  \setlength{\parskip}{-1pt}
  \setlength{\itemsep}{-1pt plus 0.1ex}
}

\title{Counterfactual Explanations for Face Forgery Detection\\ via Adversarial Removal of Artifacts}
\name{Yang~Li\textsuperscript{1,2*},
        Songlin~Yang\textsuperscript{1,2*},
        Wei~Wang\textsuperscript{1\dag},
        Ziwen~He\textsuperscript{3},
        Bo~Peng\textsuperscript{1},
        and~Jing~Dong\textsuperscript{1}
        }
\address{\textsuperscript{1}School of Artificial Intelligence, University of Chinese Academy of Sciences, China 
\\ \textsuperscript{2}CRIPAC, MAIS, Institute of Automation, Chinese Academy of Sciences, China
\\ \textsuperscript{3}Nanjing University of Information Science and Technology, China}

\begin{document}\sloppy

\def\x{{\mathbf x}}
\def\L{{\cal L}}
\definecolor{mypink}{RGB}{207, 234, 241}
\definecolor{myyellow}{RGB}{246, 202, 229}


\maketitle

    

\begin{abstract}
\makeatletter{\renewcommand*{\@makefnmark}{}
\footnotetext{$^*$ indicates equal contribution and \textsuperscript{\dag} indicates corresponding author.}}

Highly realistic AI generated face forgeries known as deepfakes have raised serious social concerns. Although DNN-based face forgery detection models have achieved good performance, they are vulnerable to latest generative methods that have less forgery traces and adversarial attacks. This limitation of generalization and robustness hinders the credibility of detection results and requires more explanations. In this work, we provide counterfactual explanations for face forgery detection from an artifact removal perspective. Specifically, we first invert the forgery images into the StyleGAN latent space, and then adversarially optimize their latent representations with the discrimination supervision from the target detection model. We verify the effectiveness of the proposed explanations from two aspects: \textbf{(1) Counterfactual Trace Visualization}: the enhanced forgery images are useful to reveal artifacts by visually contrasting the original images and two different visualization methods; \textbf{(2) Transferable Adversarial Attacks}: the adversarial forgery images generated by attacking the detection model are able to mislead other detection models, implying the removed artifacts are general. Extensive experiments demonstrate that our method achieves over 90\% attack success rate and superior attack transferability. Compared with naive adversarial noise methods, our method adopts both generative and discriminative model priors, and optimize the latent representations in a synthesis-by-analysis way, which forces the search of counterfactual explanations on the natural face manifold. Thus, more general counterfactual traces can be found and better adversarial attack transferability can be achieved. Our code is available at \small{\href{https://github.com/yangli-lab/Artifact-Eraser/}{https://github.com/yangli-lab/Artifact-Eraser/}}.

\end{abstract}

\begin{figure*}
    \centering
    \includegraphics[width=0.95\linewidth]{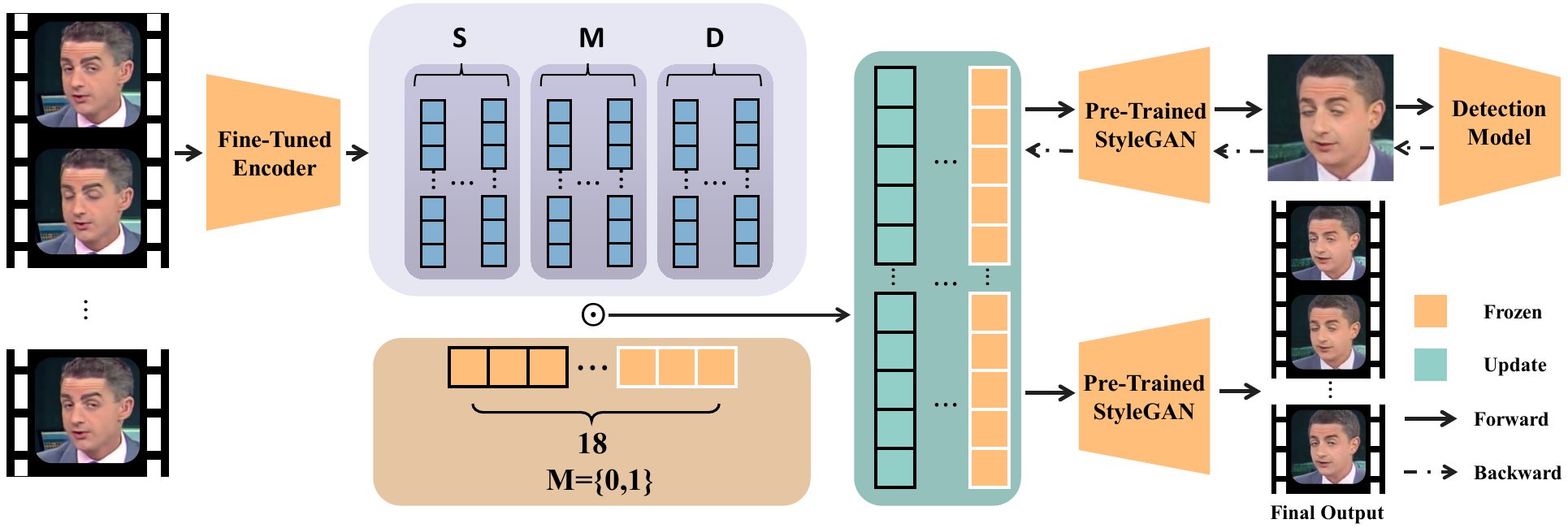}
    \caption{The overview of our method. We utilize a fine-tuned encoder and a pre-trained StyleGAN generator to optimize the latent codes of the target face forgery image or video. We generate their artifact-removed versions for counterfactual explanations.}
    \vspace{-0.5cm}
    \label{fig:pipeline}
\end{figure*}

\section{Introduction}
\label{sec:intro}

Recent advancements~\cite{zhang2022deepfake,yang2023designing,yang2023context,yang2024learning,li2024sefi} in deepfakes, especially for the face forgeries, have caused serious social concerns. To prevent the spread of malicious deepfakes, researchers have developed detection models~\cite{zhao2021multi, cao2022endtoend, proactive,zhang2018face} to discriminate whether videos or images are real or generated by synthesis techniques. However, the performance of these detectors~\cite{zhao2021multi,fox2021videoforensicshq,han2023possible} is vulnerable to latest generation methods with less forgery traces and also vulnerable to adversarial attacks~\cite{neekhara2021adversarial, yang2021systematical, yang2023exposing}, which exposes their poor generalization and robustness. To increase the credibility of these crucial detectors, explanations for detection results are important to improve and debug face forgery detection models.

Previous works use heat-maps to locate the deepfake artifacts~\cite{selvaraju2017grad,guo2023hierarchical}, but they only focus on where the traces may be, without delving into why and how the image is a forgery. As shown in Fig.~\ref{fig:teaser}, Peng et al.~\cite{peng2022counterfactual} introduced the counterfactual explanations~\cite{molnar2020interpretable} for deepfake detection, which aim to generate counterfactuals that have hypothetical realities contradicting the observed facts that are useful for interpreting detection models. However, they only focused on the face swapping fake images, and their explanations were only interpretable in the color space.

In this work, we provide a novel adversarial removal of artifact perspective to obtain counterfactual explanations for face forgery detection. We first generate the counterfactual versions of these original forgery images that are more real, which show fewer artifacts. The target artifacts include both more perceptive and less perceptive ones. Then, we provide two perspectives to verify the effectiveness of the obtained counterfactual traces: \textbf{(1) Counterfactual Trace Visualization}: the enhanced forgery images are easier for humans to notice subtle artifacts by visually contrasting the original images and two different visualization methods, and understand why these images are fake; \textbf{(2) Transferable Adversarial Attacks}: we further verify the artifacts are related to the discrimination by using the adversarial forgery images generated from one detection model to attack other detection models.

In our method, we first invert the face forgery images into the StyleGAN latent space, and then adversarially optimize their latent representations with the discrimination supervision from the target detection model. Compared with previous adversarial attacks~\cite{neekhara2021adversarial, madry2017towards, dong2018boosting}, which add noises on the images to perturb the discrimination boundaries, we optimize the adversarial perturbations in latent space. Thus, the naive black-box adversarial perturbations can be more interpretable in the synthesized results. Moreover, this synthesise-by-analysis way is able to force the search of counterfactual explanations on the natural face manifold. In this way, the more general counterfactual traces can be found and the transferable adversarial attack success rate can be improved.

\textbf{Our contributions can be summarized as follows:}
\begin{enumerate}
    \item We provide a novel counterfactual explanation for face forgery detection from an artifact removal perspective, which explains the detection results from counterfactual trace visualization and transferable adversarial attack.
    \item We optimize the latent representations in a synthesis-by-analysis way, which can find more general counterfactual traces and improve the transferable adversarial attack success rate. 
    \item Extensive experiments demonstrate that our method achieves over 90\% attack success rate and superior attack transferability across different face forgery detection models, implying the removed traces are general.
\end{enumerate}

\section{Method}
\label{sec:method}

Our objective is to generate counterfactual versions of the forgery images by eliminating the forgery traces to mislead those detectors. We achieve this by manipulating the latent vectors guided by the discriminative features obtained from target detector.
As shown in Fig.~\ref{fig:pipeline}, we first fine-tune the e4e~\cite{karras2020analyzing} encoder to improve its capability of capturing forgery traces (Sec.~\ref{subsec:finetuning}). Then, we utilize the fine-tuned encoder to extract the latent codes of forgery images. Finally, we adversarially optimize the latent representations with the discrimination supervision from the target detection model (Sec.~\ref{subsec:latent search}).

\subsection{Preliminary}
\label{subsec:preliminary}

The e4e~\cite{tov2021designing} model's latent space has shown excellent ability in encoding human face features, such as reconstruction and editing. Therefore, in this work, we utilize the e4e~\cite{tov2021designing} model to project the input deepfake into latent space. The e4e model contains an e4e encoder $\bm{E}$ and a StyleGAN~\cite{karras2020analyzing} generator $\bm{G}$. Given an image $\bm{x} \in \mathbb{R}^{H \times W \times 3}$ (where $H$ and $W$ are the image height and width), the encoder encodes it as the latent vectors $\bm{w}=\bm{E}(\bm{x})$, and then $\bm{G}$ decodes the latent vectors to get the reconstructed input $\bm{\hat{x}}=\bm{G}(\bm{E(\bm{x})}) \approx \bm{x}$. 

\subsection{Encoder Fine-Tuning}
\label{subsec:finetuning}

Let $\bm{X}$ represents a sequence of raw Deepfake video, and we denote each frame as $\bm{X}_{i}, i\in{\{1, 2, \dots, N\}}$ where $N$ is the length of the video sequence. We denote $\bm{G}$ as StyleGAN~\cite{karras2020analyzing} generator. The latent codes for each image are embedded in the extended StyleGAN space $\mathbf{\mathcal{W}} \subseteq \mathbb{R}^{k \times 512}$, where $k$ represents the number of style codes. We fine-tune the e4e~\cite{tov2021designing} encoder $\bm{E}$ to obtain the StyleGAN latent codes $\mathbf{\hat{W}}_{i}, i\in{\{1, 2, \dots, N\}}$ of $\bm{X}_{i}$ on Deepfake datasets using the following training loss:
\begin{equation}{\label{e4e_loss}}
\begin{aligned}
    \mathcal{L}_{enc}(\bm{X}, \bm{\hat{X}}) =& \lambda_{mse}\mathcal{L}_{mse}(\bm{X}, \bm{\hat{X}}) + \lambda_{lpips}\mathcal{L}_{lpips}(\bm{X}, \bm{\hat{X}}) \\
    &+ \lambda_{id}\mathcal{L}_{id}(\bm{X}, \bm{\hat{X}}),
\end{aligned}
\end{equation}
where $\bm{\hat{X}}$ is the reconstructed deepfake video sequence, and each frame is denoted by $\bm{\hat{X}}_{i}, i\in{\{1, 2, \dots, N\}}$. $\mathcal{L}_{mse}$ is the mean
square error loss between input and reconstruct image. $\mathcal{L}_{lpips}$ evaluates the perceptual similarity~\cite{zhang2018unreasonable} while $\mathcal{L}_{id}$ represents the feature embeddings extracted by ArcFace~\cite{deng2019arcface}. 


\subsection{Adversarial Searching}
\label{subsec:latent search}
\noindent
\textbf{Optimization Overview.} By utilizing the trained deepfake detectors, which are capable of capturing forgery-related features, we have developed a method that leverages these discriminative features to modify facial content with the purpose of removing the artifacts in forgery face images. We denote the target adversarial latent codes as  $\mathbf{W}^{adv}_{i}, i\in{\{1, 2, \dots, N\}}$, that can evade the face forgery detector $\bm{D}$ through subtle manipulation of the encoded latent codes $\mathbf{\hat{W}}_{i}$. Specifically, we formulate our goal as follows:
\begin{equation}{\label{attack_target}}
    \arg\min_{\mathbf{W}_{i}^{adv}} \mathcal{L}_{adv}(\bm{D}(\bm{G}(\mathbf{W}^{adv}_{i})), y_t),
\end{equation}
where $y_t$ is the target label (\textit{real} or \textit{fake}). We utilize the mean square error loss as  $\mathcal{L}_{adv}$. Our algorithm iteratively updates the vectors embedded in the latent space, and it can be expressed as follows:
\begin{equation}
        \mathbf{W}_{i}^{adv} = \mathbf{\hat{W}}_{i} + \epsilon \cdot sign(\frac{\partial \mathcal{L}_{adv}}{\partial \mathbf{\hat{W}}_{i}}).
\end{equation}

 In every iteration, we optimize the latent vectors $\mathbf{W}_{i}^{adv}=\{\bm{w}^{adv}_1, \bm{w}^{adv}_2, \dots, \bm{w}^{adv}_{18}\}$ by applying a fixed strength $\epsilon$, guided by the one-step gradient sign of loss value. 
 

\label{subsec:exlanation for Artifacts}

\noindent
\textbf{Level-Wise Strategy.} Previous studies~\cite{richardson2021encoding, li2021exploring} have demonstrated that specific style inputs of the StyleGAN latent space (e.g., $\bm{w}_1$ and $\bm{w}_{18}$) contribute to different scales of textures in the generated images. For instance, drawn from StyleGAN~\cite{karras2020analyzing}, the initial set of coarse style inputs primarily encode shape-related attributes, while the final set of fine-grained style inputs encode the detailed, microscopic features. Therefore, instead of modifying all style codes, we can selectively update the face forgery related style codes. This approach allows us to enable the attack by manipulating certain style codes while preserving some of the attributes to remain consistent with the original input.

Specifically, we categorize the latent style inputs into three levels, the shallow level (S-level) ($\bm{w}_{1}, \bm{w}_{2}, \dots, \bm{w}_{6}$) which corresponds more to the high-level features of the image, the middle level (M-level) ($\bm{w}_{7}, \bm{w}_{8}, \dots, \bm{w}_{12}$) which is more corresponding with facial features and the deep level (D-level) ($\bm{w}_{13}, \bm{w}_{14}, \dots, \bm{w}_{18}$) that primarily affects the color scheme, By dividing the style codes into these levels, we can address artifact removal in a more forgery-trace-related manner. As shown in Fig.~\ref{fig:pipeline}, we denote the mask vector as $\bm{M}\in \mathbb{R}^{18} = \{1, 0\}$, where each element determines whether a specific channel of the style codes should be updated or not, which can be formulated as:

\begin{equation}
   \mathbf{W}_{i}^{adv} = \mathbf{\hat{W}}_{i} + \bm{M}\odot(\epsilon \cdot sign(\frac{\partial \mathcal{L}_{adv}}{\partial \mathbf{\hat{W}}_{i}})).
\end{equation}

In contrast to naive adversarial attacks, we optimize the latent representations in a synthesis-by-analysis way, which can find the more general counterfactual traces and improve the transferable adversarial attack success rate. This is meaningful to the evaluation and explanation of more forgery detection models.

\section{Experiments}
\label{sec:experiments}
\begin{table*}[!th]
\footnotesize
\centering
\caption{Quality evaluation of different adversarial examples. We validate representative spatial-level adversarial attacks (i.e., adding noise on the images) and ours on frame-based detection model. \colorbox{myyellow}{Color} represents the best ID retention rate. \colorbox{mypink}{Color} represents the best values in other evaluation metrics.}
\begin{tabular}{cccccccccccccc}
    \hline
    \multirow{2}*{Model} & \multirow{2}*{Attack} & \multicolumn{4}{c}{Celeb-DF(v2)} & \multicolumn{4}{c}{DFDC} & \multicolumn{4}{c}{FF++} \\
    \cmidrule(r){3-6} \cmidrule(r){7-10} \cmidrule(r){11-14}
    ~ & ~ & ID & LPIPS$\downarrow$ & TV$\downarrow$ & ESNLE$\downarrow$ & ID& LPIPS$\downarrow$ & TV$\downarrow$ & ESNLE$\downarrow$ & ID & LPIPS$\downarrow$ & TV$\downarrow$ & ESNLE$\downarrow$ \\
    \hline\arrayrulecolor[rgb]{0.9, 0.9, 0.9}
    \multirow{4}*{Xception} & \text{FGSM} & \cellcolor{myyellow}1.00 & 0.0420 & 29.9 & 0.709 & \cellcolor{myyellow}1.00 & 0.0682 & 77.4 & 1.12 & \cellcolor{myyellow}1.00 & 0.0620 & 110 & 1.16 \\
    ~ & \text{MIFGSM} & 0.999 & 0.0440 & 31.7 & 0.731 & \cellcolor{myyellow}1.00 & 0.0722 & 78.1 & 1.13 & \cellcolor{myyellow}1.00 & 0.0640 & 111 & 1.17 \\
    ~ & $\text{PGD}{l_{inf}}$ & \cellcolor{myyellow}1.00 & 0.0420 & 29.8 & 0.709 & \cellcolor{myyellow}1.00 & 0.0681 & 77.4 & 1.12 & \cellcolor{myyellow}1.00 & 0.620 & 110 & 1.16 \\
    ~ & \text{Ours} & 0.997 & \cellcolor{mypink}0.014 & \cellcolor{mypink}18.5 & \cellcolor{mypink}0.320 & 0.998 & \cellcolor{mypink}0.0307 & \cellcolor{mypink}50.7 & \cellcolor{mypink}0.351 & 0.959 & \cellcolor{mypink}0.0404 & \cellcolor{mypink}66.7 & \cellcolor{mypink}0.395 \\
    \hline
    \multirow{4}*{Efficient-b4} & \text{FGSM} & \cellcolor{myyellow}1.00 & 0.0450 & 29.6 & 0.636 & \cellcolor{myyellow}1.00 & 0.0625 & 75.2 & 0.872 & \cellcolor{myyellow}1.00 & 0.0589 & 111 & 1.01 \\
    ~ & \text{MIFGSM} & \cellcolor{myyellow}1.00 & 0.0480 & 29.9 & 0.634 & \cellcolor{myyellow}1.00 & 0.0652 & 75.5 & 0.875 & \cellcolor{myyellow}1.00 & 0.0603 & 111 & 1.01 \\
    ~ & $\text{PGD}{l_{inf}}$ & \cellcolor{myyellow}1.00 & 0.0450 & 29.7 & 0.636 & \cellcolor{myyellow}1.00 & 0.0624 & 75.2 & 0.872 & \cellcolor{myyellow}1.00 & \cellcolor{mypink}0.0588 & 111 & 1.01 \\
    ~ & \text{Ours} & 0.999 & \cellcolor{mypink}0.0140 & \cellcolor{mypink}18.5 & \cellcolor{mypink}0.320 & 0.997 & \cellcolor{mypink}0.0448 & \cellcolor{mypink}51.2 & \cellcolor{mypink}0.365 & 0.932 & 0.0618 & \cellcolor{mypink}65.8 & \cellcolor{mypink}0.399 \\
    \arrayrulecolor[rgb]{0.9, 0.9, 0.9}\hline
    \multirow{4}*{MAT} & \text{FGSM} & \cellcolor{myyellow}1.00 & 0.0566 & 30.8 & 0.735 & \cellcolor{myyellow}1.00 & 0.0601 & 78.0 & 1.19 & \cellcolor{myyellow}1.00 & 0.0842 & 120 & 1.61 \\
    ~ & \text{MIFGSM} & \cellcolor{myyellow}1.00 & 0.0709 & 33.3 & 0.819 & 0.998 & 0.0825 & 90.8 & 1.50 & 0.999 & 0.103 & 137 & 1.85 \\
    ~ & $\text{PGD}{l_{inf}}$ & \cellcolor{myyellow}1.00 & 0.0564 & 30.8 & 0.734 & \cellcolor{myyellow}1.00 & 0.0600 & 77.9 & 1.19 & \cellcolor{myyellow}1.00 & 0.0818 & 119 & 1.58 \\
    ~ & \text{Ours} & 0.983 & \cellcolor{mypink}0.0190 & \cellcolor{mypink}18.2 & \cellcolor{mypink}0.3215 & 0.994 & \cellcolor{mypink}0.0463 & \cellcolor{mypink}48.7 & \cellcolor{mypink}0.357 & 0.912 & \cellcolor{mypink}0.0601 & \cellcolor{mypink}59.1 & \cellcolor{mypink}0.392 \\
    \arrayrulecolor[rgb]{0.9, 0.9, 0.9}\hline
    \multirow{4}*{RECCE} & \text{FGSM} & \cellcolor{myyellow}1.00 & 0.0256 & 28.4 & 0.616 & \cellcolor{myyellow}1.00 & 0.0441 & 73.9 & 0.909 & \cellcolor{myyellow}1.00 & 0.0656 & 112 & 1.22 \\
    ~ & \text{MIFGSM} & \cellcolor{myyellow}1.00 & 0.0280 & 28.8 & 0.627 & \cellcolor{myyellow}1.00 & 0.0493 & 74.9 & 0.950 & 0.999 & 0.0706 & 127 & 1.24 \\
    ~ & $\text{PGD}{l_{inf}}$ & \cellcolor{myyellow}1.00 & 0.0256 & 28.4 & 0.616 & \cellcolor{myyellow}1.00 & \cellcolor{mypink}0.0440 & 73.9 & 0.999 & 0.999 & 0.0655 & 112 & 1.22 \\
    ~ & \text{Ours}& 0.998 & \cellcolor{mypink}0.0171 & \cellcolor{mypink}18.3 & \cellcolor{mypink}0.320 & 0.992 & 0.0457 & \cellcolor{mypink}50.5 & \cellcolor{mypink}0.351 & 0.965 & \cellcolor{mypink}0.0353 & \cellcolor{mypink}68.8 & \cellcolor{mypink}0.387 \\
    \arrayrulecolor{black}\hline
    \vspace{-0.5cm}
\end{tabular}
\label{tab:quality assessment frame-based}
\end{table*}

\begin{figure}[!t]
    \vspace{-0.2cm}
    \centering
    \includegraphics[width=0.98\linewidth]{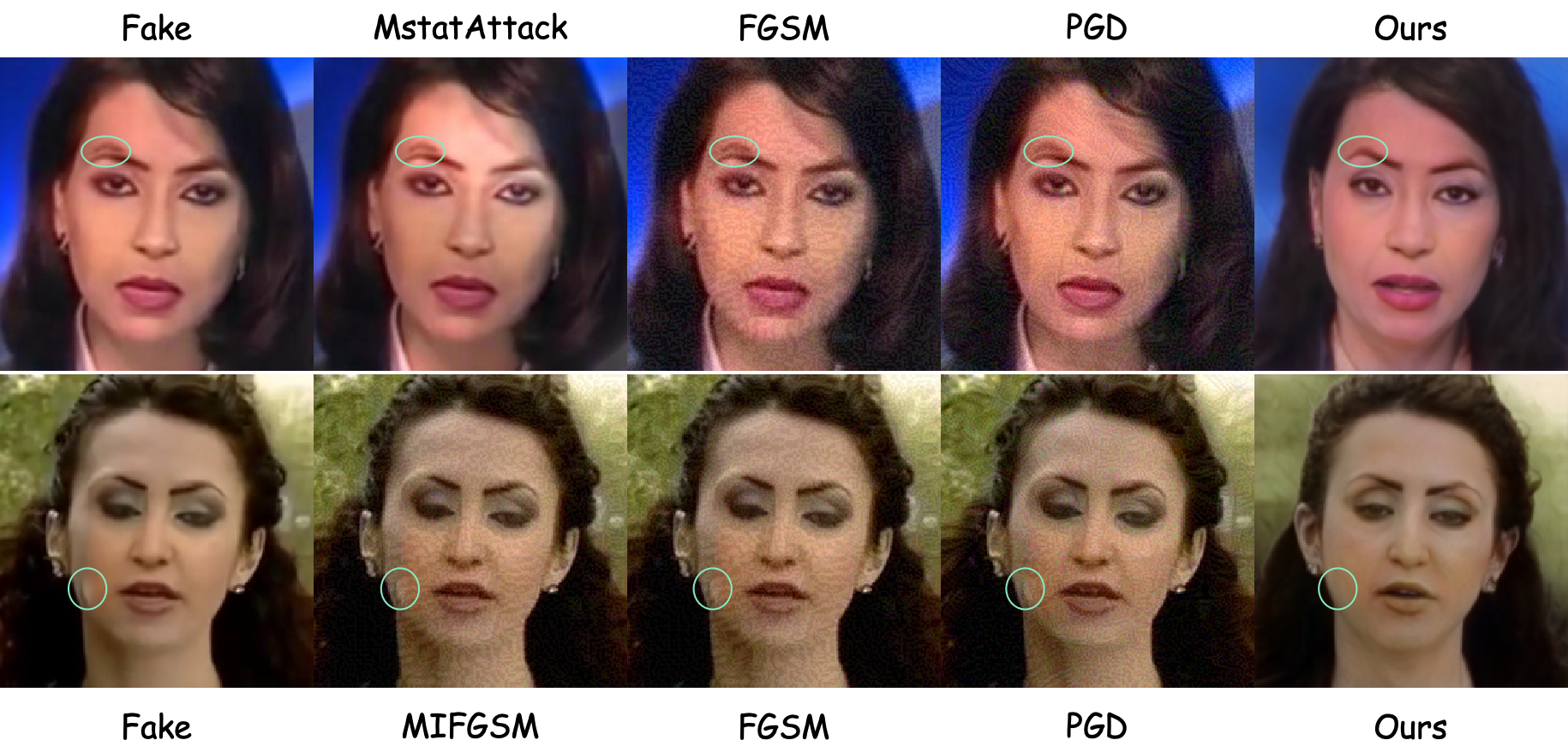}
    \caption{Visualization of the adversarial examples. The visualizations include the raw fake images and the results generated with MstatAttack~\cite{hou2023evading}, MIFGSM~\cite{dong2018boosting}, FGSM~\cite{hussain2021adversarial},  PGD~\cite{madry2017towards}, and ours. Only our proposed method can successfully remove the artifacts that we circle out.}
    \vspace{-0.5cm}
    \label{fig:fig_compare}
\end{figure}

\subsection{Experiment Setup}
\label{subsec: experiment setup}
\noindent\textbf{Datasets.} We utilize three widely used deepfake datasets, FF++~\cite{rossler2019faceforensics++}, DFDC~\cite{dolhansky2020deepfake}, and Celeb-DF(v2)~\cite{li2020celeb}. The FF++ dataset~\cite{rossler2019faceforensics++} is a popular deepfake dataset containing 1000 real videos and corresponding fake videos generated by four types of generation methods. The DFDC dataset~\cite{dolhansky2020deepfake} contains more than 100000 videos, the fake videos are created by altering the videos using a variety of different anonymous deepfake generation models. The Celeb-DF(v2) dataset~\cite{li2020celeb} contains 590 origin videos collected from YouTube and 5639 corresponding deepfake videos, which offers higher-quality deepfakes with less perceptible artifacts. For each image, we crop out the face region with face detection model~\cite{deng2019arcface}, and resize the image size to $256 \times 256$ for both training the victim models and applying the attack method on victim models.
 
\noindent\textbf{Deepfake Detectors.} In our experiments, we thoroughly evaluate the performance of our method on frame-based detectors that operate at the single-frame level. Specifically, we consider EfficientNet-b4~\cite{tan2019efficientnet}, Xception~\cite{chollet2017xception}, MAT~\cite{zhao2021multi}, RECCE~\cite{cao2022endtoend} as our target face forgery detection models.

\noindent\textbf{Baselines.} In terms of counterfactual explanations, we compare with Peng et al.~\cite{peng2022counterfactual}, EFAA~\cite{jia2022exploring}, and MstatAttack~\cite{hou2023evading}). In terms of adversarial attacks, we use the FGSM~\cite{neekhara2021adversarial} (generates adversarial samples by introducing perturbations to the input image along the gradient direction in single step), PGD~\cite{madry2017towards} (iteratively adds perturbations in pixel level), and MIFGSM~\cite{dong2018boosting} (incorporates momentum to enhance the effectiveness of perturbations) for comparisons.

\noindent\textbf{Metrics.} We use the attack success rate (ASR) as our basic evaluation metric. Additionally, we utilize both non-referenced image quality evaluation metric Total Variation (TV), noise level estimation metric ESNLE~\cite{chen2015efficient}, and full-reference image quality metric LPIPS~\cite{zhang2018unreasonable} and ID retention rate to evaluate the image quality of the generated images. We calculate ID retention rate by summarizing the ID retention rate for each test image, and we consider the two images to have same identity when the ID loss is higher than $0.75$~\cite{deng2019arcface}.

\noindent\textbf{Implementation Details.} We finetune the e4e~\cite{tov2021designing} model on the target dataset seperately. For each dataset, we set the training epochs to be 80000, with $\lambda_{id} = 0.5$, $\lambda_{lpips} = 0.8$ and $\lambda_{l_{2}} = 1.0$. We use a batch size of $8$ with a learning rate $lr = 0.0001$. After training, we keep the parameters of the models fixed during the subsequent processes. Unless otherwise stated, we set the attack strength $\epsilon$ to be $0.0006$ for Celeb-DF(v2), and $0.001$ for both DFDC and FF++ datasets.


\subsection{Counterfactual Trace Visualization}

We utilize two visualization results to provide clear and thorough counterfactual explanations for face forgery traces: discrimination activation map (Grad-CAM heat-maps~\cite{selvaraju2017grad}) and residual maps. As shown in Fig.~\ref{fig:teaser}, we compare our approach with Peng et al.~\cite{peng2022counterfactual} and our method offers broader applicability by targeting artifact removal for more various forgeries, instead of relying on a face mask to specify the artifact location of face swapping regions. 

Different from those spatial-level adversarial perturbations, which struggle to locate the artifact areas, our manipulation of latent space is semantically interpretable, such as asymmetry eyes and illumination inconsistency, as shown in Fig.~\ref{fig:teaser}. Moreover, previous adversarial methods mislead the detectors by adding noise on images (e.g., FGSM~\cite{neekhara2021adversarial}, MIFGSM~\cite{dong2018boosting}, and PGD~\cite{madry2017towards}) or certain shaded areas (e.g., MStatAttack~\cite{hou2023evading}) and less attention is paid to the forgery traces in the face regions, as shown in Fig.~\ref{fig:fig_compare}. Consequently, these adversarial examples remain forgery traces and can still be detected by other detectors. As depicted in the blue block in Fig.~\ref{fig:teaser}, except for the more human-eye-sensitive artifacts, deep generative methods can generate deepfakes that can be detected by models but are less human-eye-sensitive. With our proposed artifact removal method, both more visible and less visible face forgery traces can be visualized clearly, including their location, shape and the color differences. Overall, our method provides a more human-friendly and clear visualization results for face forgery traces.

\subsection{Transferable Adversarial Attacks}

Besides visualization, our method verifies the effectiveness of the proposed explanations from transferable adversarial attacks, implying the removed artifacts are general. We validate our attack performance from quality and transferability, respectively. 

\noindent
\textbf{Quality.} As shown in Tab.~\ref{tab:quality assessment frame-based}, we compare image quality of our generated examples with baselines on the images that successfully evade the target detectors. Our proposed method demonstrates its effectiveness in concealing forgery traces while preserving the face identity. Moreover, the TV, LPIPS, and ESNLE scores suggest the superior image quality of our method. Specifically, the TV and ESNLE scores are approximately $50\%$ lower compared to those of FGSM, MIFGSM, and PGD.

\noindent
\textbf{Transferability.} As shown in Tab.~\ref{tab:transferability celebdf}, the results on Celeb-DF(v2)~\cite{li2020celeb} dataset indicate that the generated images exhibit higher transferability than the others. For instance, the examples generated by RECCE with our method get $60\%$ higher success rate than FGSM~\cite{neekhara2021adversarial} and PGD~\cite{madry2017towards}. Moreover, we also present the performance on FF++~\cite{rossler2019faceforensics++} dataset in Tab.~\ref{tab:transferability ffpp}. The PGD~\cite{madry2017towards}, FGSM~\cite{neekhara2021adversarial, hussain2021adversarial}, and the SOTA method EFAA~\cite{jia2022exploring} are used as comparisons. The images generated with Efficient-b4 achieve $30\%$ higher transferability than the others on Xception. Since deepfake images in Celeb-DF(v2)~\cite{li2020celeb} have better visual quality, it would be relatively easier for us to remove forgery traces. Instead, the data in FF++~\cite{rossler2019faceforensics++} is more complex and of low quality, making it challenging to remove artifact content. Consequently, this leads to lower transferability scores. Furthermore, latent searching faces inherent challenges in realizing fine-grained semantic feature manipulations, leading to a comparatively diminished attack success rate on the targeted model in comparison to the other methods. Similar trends can be observed in DFDC~\cite{dolhansky2020deepfake}, please refer to the supplementary for the details.
\begin{table}[!th]
\scriptsize
    \centering
    \caption{The transferability of the proposed artifact removal along with other comparison methods on Celeb-DF(v2)~\cite{li2020celeb} dataset. \textbf{Bold} indicates the highest attack success rate. Other detection models find it challenging to detect the artifacts removed samples.}
    \begin{tabular}{cccccc}
    \hline
    Model & Attack & Efficient-b4 & Xception & MAT & RECCE \\
    \hline
        \multirow{4}*{Efficient-b4} & FGSM & 99.2 & 17.2 & 30.3 & 12.3 \\
        ~ & $\text{PGD}{l_{inf}}$& \textbf{99.9} & 18.5 & 31.2 & 12.2 \\
        ~ & Ours & 98.9 & \textbf{85.7} & \textbf{75.8} & \textbf{54.0}\\
    \arrayrulecolor[rgb]{0.9, 0.9, 0.9}\hline
        \multirow{4}*{Xception} & FGSM & 4.36 & 99.3 & 7.15 & 26.5 \\
        ~ & $\text{PGD}{l_{inf}}$ & 4.86 & \textbf{100} & 7.15 & 26.5 \\
        ~ & Ours & \textbf{86.1} & 99.2 & \textbf{73.9} & \textbf{73.0} \\
    \hline
        \multirow{4}*{MAT} & FGSM & 18.0 & 21.4 & 99.8 & 32.9 \\
        ~ & $\text{PGD}{l_{inf}}$ & 18.0 & 21.4 & \textbf{100} & 32.8 \\
        ~ & Ours & \textbf{80.6} & \textbf{82.0} & 90.6 & \textbf{70.0} \\
    \hline
        \multirow{4}*{RECCE} & FGSM & 9.74 & 26.0 & 26.7 & 99.5 \\
        ~ & $\text{PGD}{l_{inf}}$ & 9.62 & 26.0 & 28.2 & \textbf{100} \\
        ~ & Ours & \textbf{89.6} & \textbf{95.0} & \textbf{90.0} & 98.8 \\
    \arrayrulecolor{black}\hline
    \end{tabular}
    \label{tab:transferability celebdf}
    \vspace{-0.2cm}
\end{table}

\begin{table}[!th]
\scriptsize
    \centering
    \caption{The transferability of the proposed artifact removal along with other comparison methods on FF++~\cite{rossler2019faceforensics++} dataset. * represents results taken from~\cite{jia2022exploring}. Other detection models find it challenging to detect the artifacts removed samples.}
    \begin{tabular}{cccccc}
    \hline
    Model & Attack & Efficient-b4 & Xception & MAT & RECCE \\
    \hline
        \multirow{4}*{Efficient-b4} & FGSM & 38.7* & 0.9* & 1.2 & 1.0 \\
        ~ & $\text{PGD}{l_{inf}}$& 71.6* & 0.3* & 1.3 & 0.9 \\
        ~ & EFAA~\cite{jia2022exploring}& \textbf{83.2*} & 1.4* & / & / \\
        ~ & Ours & 67.9 & \textbf{40.2} & \textbf{52.6} & \textbf{39.8}\\
    \arrayrulecolor[rgb]{0.9, 0.9, 0.9}\hline
        \multirow{4}*{Xception} & FGSM & 1.1* & 18.9* & 0.9 & 2.1 \\
        ~ & $\text{PGD}{l_{inf}}$ & 1.1* & 61.6* & 1.0 & 1.8 \\
        ~ & EFAA~\cite{jia2022exploring} & 1.5* & \textbf{70.5*} & / & / \\
        ~ & Ours & \textbf{11.4} & 60.2 & \textbf{12.5} & \textbf{44.4} \\
    \hline 
        \multirow{4}*{MAT} & FGSM & 2.1 & 1.2 & 52.3 & 1.6 \\
        ~ & $\text{PGD}{l_{inf}}$ & 2.3 & 1.4 & \textbf{80.2} & 1.4 \\
        ~ & EFAA~\cite{jia2022exploring}& / & / & / & / \\
        ~ & Ours & \textbf{39.5} & \textbf{38.2} & 79.6 & \textbf{36.5} \\
    \hline
        \multirow{4}*{RECCE} & FGSM & 1.9 & 2.5 & 2.1 & 62.1 \\
        ~ & $\text{PGD}{l_{inf}}$ & 2.1 & 2.3 & 1.8 & \textbf{81.3} \\
        ~ & EFAA~\cite{jia2022exploring} & / & / & / & / \\
        ~ & Ours & \textbf{12.1} & \textbf{48.2} & \textbf{13.2} & 60.8 \\
    \arrayrulecolor{black}\hline
    \end{tabular}
    \label{tab:transferability ffpp}
    \vspace{-0.2cm}
\end{table}

\begin{table}[!th]
\scriptsize
    \centering
    \caption{The ASR under different level-wise strategies. We evaluate on EfficientNet-b4~\cite{tan2019efficientnet} with a fixed query number $100$. The results suggest that the M-level style codes are more relevant to forgery features in deepfake images.}
    \begin{tabular}{ccccc}
    \hline
       Dataset  & S-level & M-level & D-level & Full \\
    \hline
       FF++($\epsilon=0.0012$) & 24.11 & 60.05 & 23.25 & \textbf{91.80} \\
       DFDC($\epsilon=0.0008$) & 11.78 & 41.28 & 13.01 & \textbf{73.17}\\
       Celeb-DF(v2)($\epsilon=0.0004$) & 30.95 & 67.33 & 30.55 & \textbf{93.34}\\
    \hline
    \end{tabular}
    \label{tab:masked attacking suc rate}
    \vspace{-0.3cm}
\end{table}

\subsection{Ablation Study}
\label{subsec: ablation study}

\noindent\textbf{Level-Wise Strategy.} There exists human-eye-sensitive artifacts (D-level) and visible distortion (S-level), and subtle facial artifacts (M-level) that are not human-eye-sensitive, like minor distortions, irregular eyebrow thickness~\cite{peng2022counterfactual} and asymmetric eyes. To further explain the composition of these artifacts in deepfake images, we try to differentiate the common forgery representations by comparing the attack success rate of selectively masking different groups of style vectors: S-level, M-level, and D-level. During each iteration, we only update the specific six style input vectors and freeze the masked latent codes. Tab.~\ref{tab:masked attacking suc rate} shows the ASR of different mask setting. The findings in Tab.~\ref{tab:masked attacking suc rate} suggest that the facial abnormal minor artifacts are the most commonly discriminative features that persist during the deepfake generation process.

\noindent\textbf{Ablation on Attack Strength $\epsilon$.} According to the results in Table~\ref{tab:attack strength}, though raising in attack success rate, increasing $\epsilon$ beyond a certain threshold may lead to a subtle degradation in the quality of the generated adversarial samples, which may interfere the visualization of artifact traces.

\begin{table}[th]
\footnotesize
    \centering
    \caption{The trade-off between attacking strength and image quality. We conduct the experiments on FF++ dataset with fixed loop size $100$.}
     \begin{tabular}{c|ccccc}
     \hline
     Victim & $\epsilon$ & ID$\uparrow$ & LPIPS$\downarrow$ & ESNLE$\downarrow$ & ASR \\
     \hline
        \multirow{4}*{Xception} & 0.0006 & \textbf{0.973} & \textbf{0.0362} & \textbf{0.345} & 57.3 \\
        ~ & 0.0008 & 0.972 & 0.0382 & 0.377 & 58.4 \\
        ~ & 0.001 & 0.965 & 0.0398 & 0.382 & 60.2 \\
        ~ & 0.0012 & 0.951 & 0.0421 & 0.404 & \textbf{65.2} \\
    \cline{1-6}
        \multirow{4}*{Efficient-b4} & 0.0006 & \textbf{0.969} & \textbf{0.0537} & \textbf{0.372} & 58.9 \\
        ~ & 0.0008 & 0.955 & 0.0578 & 0.385 & 65.3 \\
        ~ & 0.001 & 0.943 & 0.0603 & 0.389 & 67.9 \\
        ~ & 0.0012 & 0.930 & 0.0623 & 0.417  & \textbf{71.4} \\
    \hline
    \end{tabular}
    \vspace{-0.6cm}
    \label{tab:attack strength}
\end{table}
\section{Conclusion}
\label{sec:conclusion}
\vspace{-0.2cm}
In this work, we provide counterfactual explanations for face forgery detection by adversarially removing artifacts, we validate the effectiveness of our proposed explanations from two perspectives: counterfactual trace visualization and transferable adversarial attacks. Extensive experiments demonstrate that our method achieves over 90\% attack success rate and superior attack transferability across various face forgery detection models, implying the artifacts removed by our method possess a general nature.

\noindent\textbf{Acknowledgments:} This work is supported by the National Natural Science Foundation of China (NSFC) under Grants 62372452, 62272460.

\scriptsize
\bibliographystyle{IEEEbib}
\bibliography{icme2023template}

\begin{thebibliography}{10}

\bibitem{peng2022counterfactual}
Bo~Peng, Siwei Lyu, Wei Wang, and Jing Dong,
\newblock ``Counterfactual image enhancement for explanation of face swap deepfakes,''
\newblock in {\em Chinese Conference on Pattern Recognition and Computer Vision (PRCV)}. Springer, 2022, pp. 492--508.

\bibitem{selvaraju2017grad}
Ramprasaath~R Selvaraju, Michael Cogswell, Abhishek Das, Ramakrishna Vedantam, Devi Parikh, and Dhruv Batra,
\newblock ``Grad-cam: Visual explanations from deep networks via gradient-based localization,''
\newblock in {\em Proceedings of the IEEE international conference on computer vision}, 2017, pp. 618--626.

\bibitem{zhang2022deepfake}
Tao Zhang,
\newblock ``Deepfake generation and detection, a survey,''
\newblock {\em Multimedia Tools and Applications}, vol. 81, no. 5, pp. 6259--6276, 2022.

\bibitem{yang2023designing}
Songlin Yang, Wei Wang, Bo~Peng, and Jing Dong,
\newblock ``Designing a 3d-aware stylenerf encoder for face editing,''
\newblock in {\em ICASSP 2023-2023 IEEE International Conference on Acoustics, Speech and Signal Processing (ICASSP)}. IEEE, 2023, pp. 1--5.

\bibitem{yang2023context}
Songlin Yang, Wei Wang, Jun Ling, Bo~Peng, Xu~Tan, and Jing Dong,
\newblock ``Context-aware talking-head video editing,''
\newblock in {\em Proceedings of the 31st ACM International Conference on Multimedia}, 2023, pp. 7718--7727.

\bibitem{yang2024learning}
Songlin Yang, Wei Wang, Yushi Lan, Xiangyu Fan, Bo~Peng, Lei Yang, and Jing Dong,
\newblock ``Learning dense correspondence for nerf-based face reenactment,''
\newblock in {\em Proceedings of the AAAI Conference on Artificial Intelligence}, 2024, vol.~38, pp. 6522--6530.

\bibitem{li2024sefi}
Yang Li, Songlin Yang, Wei Wang, and Jing Dong,
\newblock ``Beyond inserting: Learning identity embedding for semantic-fidelity personalized diffusion generation,''
\newblock {\em arXiv preprint arXiv:2402.00631}, 2024.

\bibitem{zhao2021multi}
Hanqing Zhao, Wenbo Zhou, Dongdong Chen, Tianyi Wei, Weiming Zhang, and Nenghai Yu,
\newblock ``Multi-attentional deepfake detection,''
\newblock in {\em Proceedings of the IEEE/CVF conference on computer vision and pattern recognition}, 2021, pp. 2185--2194.

\bibitem{cao2022endtoend}
Junyi Cao, Chao Ma, Taiping Yao, Shen Chen, Shouhong Ding, and Xiaokang Yang,
\newblock ``End-to-end reconstruction-classification learning for face forgery detection,''
\newblock in {\em Proceedings of the IEEE/CVF Conference on Computer Vision and Pattern Recognition (CVPR)}, June 2022, pp. 4113--4122.

\bibitem{proactive}
Yuan Zhao, Bo~Liu, Ming Ding, Baoping Liu, Tianqing Zhu, and Xin Yu,
\newblock ``Proactive deepfake defence via identity watermarking,''
\newblock in {\em Proceedings of the IEEE/CVF Winter Conference on Applications of Computer Vision (WACV)}, January 2023, pp. 4602--4611.

\bibitem{zhang2018face}
Le-Bing Zhang, Fei Peng, and Min Long,
\newblock ``Face morphing detection using fourier spectrum of sensor pattern noise,''
\newblock in {\em 2018 IEEE international conference on multimedia and expo (ICME)}. IEEE, 2018, pp. 1--6.

\bibitem{fox2021videoforensicshq}
Gereon Fox, Wentao Liu, Hyeongwoo Kim, Hans-Peter Seidel, Mohamed Elgharib, and Christian Theobalt,
\newblock ``Videoforensicshq: Detecting high-quality manipulated face videos,''
\newblock in {\em 2021 IEEE International Conference on Multimedia and Expo (ICME)}. IEEE, 2021, pp. 1--6.

\bibitem{han2023possible}
Xiaoxuan Han, Songlin Yang, Wei Wang, Ziwen He, and Jing Dong,
\newblock ``Is it possible to backdoor face forgery detection with natural triggers?,''
\newblock {\em arXiv preprint arXiv:2401.00414}, 2023.

\bibitem{neekhara2021adversarial}
Paarth Neekhara, Brian Dolhansky, Joanna Bitton, and Cristian~Canton Ferrer,
\newblock ``Adversarial threats to deepfake detection: A practical perspective,''
\newblock in {\em Proceedings of the IEEE/CVF conference on computer vision and pattern recognition}, 2021, pp. 923--932.

\bibitem{yang2021systematical}
Songlin Yang, Wei Wang, Yuehua Cheng, and Jing Dong,
\newblock ``A systematical solution for face de-identification,''
\newblock in {\em Biometric Recognition: 15th Chinese Conference, CCBR 2021, Shanghai, China, September 10--12, 2021, Proceedings 15}. Springer, 2021, pp. 20--30.

\bibitem{yang2023exposing}
Songlin Yang, Wei Wang, Chenye Xu, Ziwen He, Bo~Peng, and Jing Dong,
\newblock ``Exposing fine-grained adversarial vulnerability of face anti-spoofing models,''
\newblock in {\em Proceedings of the IEEE/CVF Conference on Computer Vision and Pattern Recognition}, 2023, pp. 1001--1010.

\bibitem{guo2023hierarchical}
Xiao Guo, Xiaohong Liu, Zhiyuan Ren, Steven Grosz, Iacopo Masi, and Xiaoming Liu,
\newblock ``Hierarchical fine-grained image forgery detection and localization,''
\newblock in {\em Proceedings of the IEEE/CVF Conference on Computer Vision and Pattern Recognition}, 2023, pp. 3155--3165.

\bibitem{molnar2020interpretable}
Christoph Molnar,
\newblock {\em Interpretable machine learning},
\newblock Lulu. com, 2020.

\bibitem{madry2017towards}
Aleksander Madry, Aleksandar Makelov, Ludwig Schmidt, Dimitris Tsipras, and Adrian Vladu,
\newblock ``Towards deep learning models resistant to adversarial attacks,''
\newblock {\em arXiv preprint arXiv:1706.06083}, 2017.

\bibitem{dong2018boosting}
Yinpeng Dong, Fangzhou Liao, Tianyu Pang, Hang Su, Jun Zhu, Xiaolin Hu, and Jianguo Li,
\newblock ``Boosting adversarial attacks with momentum,''
\newblock in {\em Proceedings of the IEEE conference on computer vision and pattern recognition}, 2018, pp. 9185--9193.

\bibitem{karras2020analyzing}
Tero Karras, Samuli Laine, Miika Aittala, Janne Hellsten, Jaakko Lehtinen, and Timo Aila,
\newblock ``Analyzing and improving the image quality of stylegan,''
\newblock in {\em Proceedings of the IEEE/CVF conference on computer vision and pattern recognition}, 2020, pp. 8110--8119.

\bibitem{tov2021designing}
Omer Tov, Yuval Alaluf, Yotam Nitzan, Or~Patashnik, and Daniel Cohen-Or,
\newblock ``Designing an encoder for stylegan image manipulation,''
\newblock {\em ACM Transactions on Graphics (TOG)}, vol. 40, no. 4, pp. 1--14, 2021.

\bibitem{zhang2018unreasonable}
Richard Zhang, Phillip Isola, Alexei~A Efros, Eli Shechtman, and Oliver Wang,
\newblock ``The unreasonable effectiveness of deep features as a perceptual metric,''
\newblock in {\em Proceedings of the IEEE conference on computer vision and pattern recognition}, 2018, pp. 586--595.

\bibitem{deng2019arcface}
Jiankang Deng, Jia Guo, Niannan Xue, and Stefanos Zafeiriou,
\newblock ``Arcface: Additive angular margin loss for deep face recognition,''
\newblock in {\em Proceedings of the IEEE/CVF Conference on Computer Vision and Pattern Recognition (CVPR)}, June 2019.

\bibitem{richardson2021encoding}
Elad Richardson, Yuval Alaluf, Or~Patashnik, Yotam Nitzan, Yaniv Azar, Stav Shapiro, and Daniel Cohen-Or,
\newblock ``Encoding in style: a stylegan encoder for image-to-image translation,''
\newblock in {\em Proceedings of the IEEE/CVF conference on computer vision and pattern recognition}, 2021, pp. 2287--2296.

\bibitem{li2021exploring}
Dongze Li, Wei Wang, Hongxing Fan, and Jing Dong,
\newblock ``Exploring adversarial fake images on face manifold,''
\newblock in {\em Proceedings of the IEEE/CVF Conference on Computer Vision and Pattern Recognition}, 2021, pp. 5789--5798.

\bibitem{hou2023evading}
Yang Hou, Qing Guo, Yihao Huang, Xiaofei Xie, Lei Ma, and Jianjun Zhao,
\newblock ``Evading deepfake detectors via adversarial statistical consistency,''
\newblock in {\em Proceedings of the IEEE/CVF Conference on Computer Vision and Pattern Recognition}, 2023, pp. 12271--12280.

\bibitem{hussain2021adversarial}
Shehzeen Hussain, Paarth Neekhara, Malhar Jere, Farinaz Koushanfar, and Julian McAuley,
\newblock ``Adversarial deepfakes: Evaluating vulnerability of deepfake detectors to adversarial examples,''
\newblock in {\em Proceedings of the IEEE/CVF winter conference on applications of computer vision}, 2021, pp. 3348--3357.

\bibitem{rossler2019faceforensics++}
Andreas Rossler, Davide Cozzolino, Luisa Verdoliva, Christian Riess, Justus Thies, and Matthias Nie{\ss}ner,
\newblock ``Faceforensics++: Learning to detect manipulated facial images,''
\newblock in {\em Proceedings of the IEEE/CVF international conference on computer vision}, 2019, pp. 1--11.

\bibitem{dolhansky2020deepfake}
Brian Dolhansky, Joanna Bitton, Ben Pflaum, Jikuo Lu, Russ Howes, Menglin Wang, and Cristian~Canton Ferrer,
\newblock ``The deepfake detection challenge (dfdc) dataset,''
\newblock {\em arXiv preprint arXiv:2006.07397}, 2020.

\bibitem{li2020celeb}
Yuezun Li, Xin Yang, Pu~Sun, Honggang Qi, and Siwei Lyu,
\newblock ``Celeb-df: A large-scale challenging dataset for deepfake forensics,''
\newblock in {\em Proceedings of the IEEE/CVF conference on computer vision and pattern recognition}, 2020, pp. 3207--3216.

\bibitem{tan2019efficientnet}
Mingxing Tan and Quoc Le,
\newblock ``Efficientnet: Rethinking model scaling for convolutional neural networks,''
\newblock in {\em International conference on machine learning}. PMLR, 2019, pp. 6105--6114.

\bibitem{chollet2017xception}
Fran{\c{c}}ois Chollet,
\newblock ``Xception: Deep learning with depthwise separable convolutions,''
\newblock in {\em Proceedings of the IEEE conference on computer vision and pattern recognition}, 2017, pp. 1251--1258.

\bibitem{jia2022exploring}
Shuai Jia, Chao Ma, Taiping Yao, Bangjie Yin, Shouhong Ding, and Xiaokang Yang,
\newblock ``Exploring frequency adversarial attacks for face forgery detection,''
\newblock in {\em Proceedings of the IEEE/CVF Conference on Computer Vision and Pattern Recognition}, 2022, pp. 4103--4112.

\bibitem{chen2015efficient}
Guangyong Chen, Fengyuan Zhu, and Pheng Ann~Heng,
\newblock ``An efficient statistical method for image noise level estimation,''
\newblock in {\em Proceedings of the IEEE International Conference on Computer Vision}, 2015, pp. 477--485.

\end{thebibliography}

\appendix
\section{Additional Results}

\subsection{Additional Quantitative Results}

\noindent\textbf{Transferable Adversarial Attacks on DFDC~\cite{dolhansky2020deepfake} Dataset.} We evaluate the transferability on DFDC dataset, the results are listed in Tab.~\ref{tab:transferability dfdc}. Adversarial samples generated by our method on RECCE~\cite{cao2022endtoend} achieve $20\%$ higher ASR on MAT~\cite{zhao2021multi} than FGSM~\cite{hussain2021adversarial}, MIFGSM~\cite{dong2018boosting}, PGD~\cite{madry2017towards}, which demonstrates the effectiveness of our method in removing artifacts.
\begin{figure}[!th]
    \centering
    \includegraphics[width=0.98\linewidth]{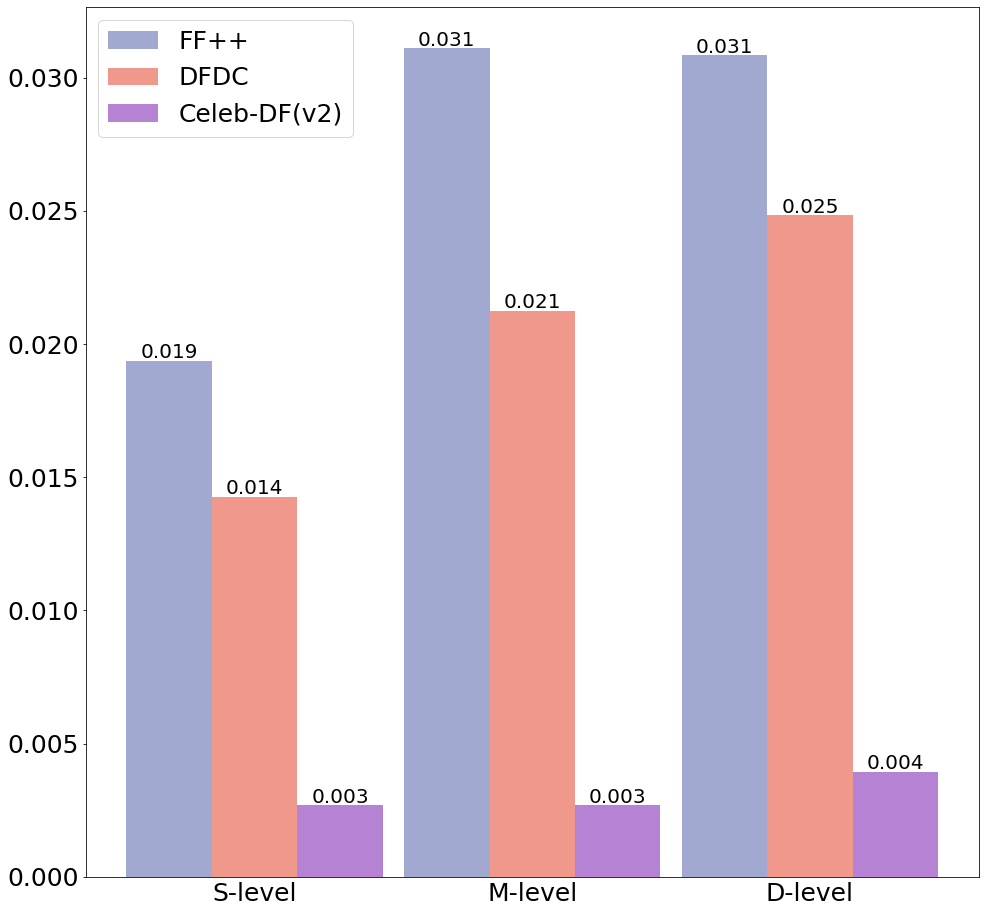}
    \caption{Changes of latent codes observed in images that successfully evade the detectors.}
    \label{fig:sup_compare_dataset2}
\end{figure}

\begin{table}[!ht]
\scriptsize
    \centering
    \caption{The transferability of the proposed artifact removal along with other comparison methods on DFDC~\cite{dolhansky2020deepfake} dataset. \textbf{Bold} indicates the highest attack success rate. Examples generated by our method exhibit better transferability across models.}
    \begin{tabular}{cccccc}
    \hline
    Victim & Attack & Efficient & Xception & MAT & RECCE \\
    \hline
        \multirow{4}*{{Efficient}} & FGSM & 87.9 & 0.63 & 1.97 & 0.051 \\
        ~ & MIFGSM & 96.8 & 0.72 & 2.07 & 0.12 \\
        ~ & $\text{PGD}{l_{inf}}$& 96.9 & 0.85 & 1.52 & 0.21 \\
        ~ & Ours & 89.9 & \textbf{13.9} & \textbf{14.4} & \textbf{0.54}\\
        \arrayrulecolor[rgb]{0.9, 0.9, 0.9}\hline
        \multirow{4}*{Xception} & FGSM & 2.91 & 86.9 & 0.85 & 4.8 \\
        ~ & MIFGSM & 2.36 & 98.6 & 0.65 & 4.9 \\
        ~ & $\text{PGD}{l_{inf}}$ & 2.86 & 99.0 & 1.15 & 3.5 \\
        ~ & Ours & \textbf{13.4} & 92.4 & \textbf{13.2} & \textbf{14.9} \\
        \hline
        \multirow{4}*{MAT} & FGSM & 4.74 & 1.98 & 84.2 & 2.27 \\
        ~ & MIFGSM & 4.12 & 2.15 & 90.22 & 2.01 \\
        ~ & $\text{PGD}{l_{inf}}$ & 3.10 & 1.98 & 91.2 & 1.48 \\
        ~ & Ours & \textbf{18.0} & \textbf{17.0} & 80.6 & \textbf{3.49} \\
        \hline
        \multirow{4}*{RECCE} & FGSM & 1.90 & 1.40 & 1.46 & 79.9 \\
        ~ & MIFGSM & 1.22 & 1.12 & 0.86 & 84.5 \\
        ~ & $\text{PGD}{l_{inf}}$ & 0.75 & 1.10 & 1.32 & 88.4 \\
        ~ & Ours & \textbf{11.0} & \textbf{15.57} & \textbf{22.7} & 86.0 \\
    \arrayrulecolor{black}\hline
    \end{tabular}
    \label{tab:transferability dfdc}
\end{table}

\noindent\textbf{Modification Strength Evaluation.} In Fig.~\ref{fig:sup_compare_dataset2}, we present a comparison of the changes in latent codes observed in images that successfully evade the detectors across different datasets and groups of style vectors. The results demonstrate that it requires less effort to remove forgery content of images from Celeb-DF(v2)~\cite{li2020celeb} than DFDC~\cite{dolhansky2020deepfake} and FF++~\cite{rossler2019faceforensics++}. The increased complexity in erasing artifact content in DFDC and FF++ images can be attributed to various factors. Firstly, the images in these datasets are complex and of lower quality, which makes the erasing process hard to be carried out. Secondly, there might be confusion in the gradient guidance provided by the detectors. This confusion could restrict the modifications to the most relevant part of the forgery traces detected by the test detector in the images. Consequently, these modifications may not effectively address other artifacts that are present in the images. Those limitations contribute to the lower transferability observed in FF++ and DFDC datasets.

\subsection{Additional Qualitative Results}
\begin{figure*}[ht]
    \centering
    \includegraphics[width=0.95\linewidth]{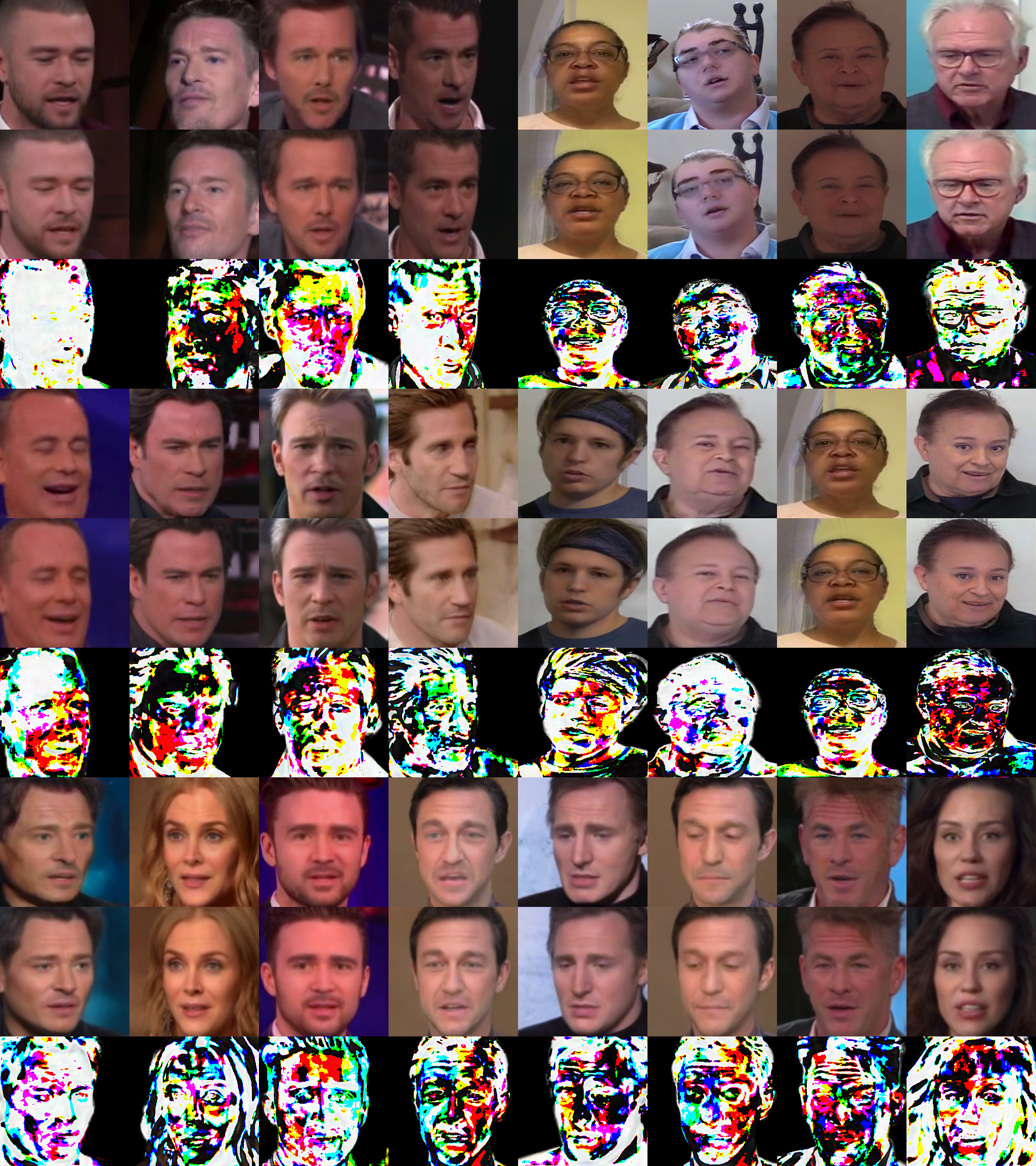}
    \caption{Visualization results of our method.}
    \label{fig:more_attacked_result}
\end{figure*}

Fig.~\ref{fig:more_attacked_result} shows more examples generated by our method, our artifacts removal can successfully localize the artifacts in naive \textit{fake} images and conceal them. 

\section{Implementation Details}

\subsection{Training for Inversion}
We finetune the e4e~\cite{tov2021designing} model on the target dataset seperately. For each dataset, we set the training epochs to 80000, with $\lambda_{id} = 0.5$, $\lambda_{lpips} = 0.8$, and $\lambda_{l_{2}} = 1.0$. We use a batch size of $8$ with a learning rate $lr = 0.0001$. During training, we freeze the parameters of the decoder, and do not employ the progressive training strategy used in origin paper~\cite{tov2021designing}. After training, we keep the parameters of the models fixed during subsequent processes.

\subsection{Basic Setting}

\noindent\textbf{The DFDC~\cite{dolhansky2020deepfake} Dataset.} The DFDC dataset is a challenging dataset consisting more than 100,000 videos, the fake videos are created by altering the videos using a variety of different anonymous Deepfake generation models. For our evaluation, we specifically utilize the $dfdc\_train\_part\_0$ and $dfdc\_train\_part\_49$ as our test data.

\noindent\textbf{Traditional Norm-Based Attack Method.} To make a detailed comparison with the norm-based attack method, we compare with the commonly used white-box attacking algorithms FGSM~\cite{hussain2021adversarial}, MIFGSM~\cite{dong2018boosting}, and $\text{PGD}{l_{inf}}$~\cite{madry2017towards}. For each dataset, we set the attacking strength $\epsilon$ fixed, specifically $0.007$ for Celeb-DF(v2), $0.011$ for DFDC and $0.015$ for FF++ in order to construct adversarial examples more effectively. Moreover, we set the attacking boundary $\beta$ for all images to be $0.1$.

\noindent\textbf{Ours.} Unless otherwise stated, we set the attack strength $\epsilon$ to be $0.0006$ for Celeb-DF(v2), and $0.001$ for both DFDC and FF++ datasets.

\end{document}